\documentclass{article}

\usepackage[preprint]{nips_2018}

\usepackage[utf8]{inputenc} 
\usepackage[T1]{fontenc}    
\usepackage{hyperref}       
\usepackage{url}            
\usepackage{booktabs}       
\usepackage{amsfonts}

\newtheorem{lemma}{Lemma}       
\newtheorem{definition}{Definition}
\usepackage{nicefrac}       
\usepackage{microtype}      
\usepackage{subfigure}
\usepackage{graphicx}
 \usepackage{mathrsfs}
\usepackage{float}
\usepackage{multirow}
\usepackage{amsmath}
\title{Knowledge-based Fully Convolutional Network and Its
Application in Segmentation of Lung CT Images}

\author{
  Tao Yu\thanks{First author} \\
  Department of Mathematics\\
  Shanghai Jiao Tong University\\
  Shanghai, China \\
  \texttt{ydtydr@sjtu.edu.cn} \\
  \And
  Yu Qiao\thanks{Corresponding author}\\
  Department of Automation\\
  Shanghai Jiao Tong University\\
  Shanghai, China \\
  \texttt{qiaoyu@sjtu.edu.cn} \\
  \And
  Huan Long\thanks{Coauthor}\\
  Department of Computer Science and Engineering\\
  Shanghai Jiao Tong University\\
  Shanghai, China \\
  \texttt{longhuan@cs.sjtu.edu.cn} \\}

\begin{document}

\maketitle

\begin{abstract}
  A variety of deep neural networks have been applied in medical
  image segmentation and achieve good performance. Unlike natural
  images, medical images of the same imaging modality are characterized
  by the same pattern, which indicates that same normal organs or tissues
  locate at similar positions in the images. Thus, in this paper we
  try to incorporate the prior knowledge of medical images
into the structure of neural networks such that the prior knowledge
can be utilized for accurate segmentation. Based on this
idea, we propose a novel deep network called knowledge-based fully
convolutional network (KFCN) for medical image segmentation.
The segmentation function and corresponding error is analyzed.
We show the existence of an asymptotically stable region for KFCN
which traditional FCN doesn't possess.  Experiments validate our knowledge assumption
about the incorporation of prior knowledge into the convolution kernels of KFCN
and show that KFCN can achieve a reasonable segmentation and a satisfactory
 accuracy.
\end{abstract}

\section{Introduction}
Medical image segmentation is of key importance for computer-aided
diagnosis system. Accurate segmentation of healthy tissues and suspicious
lesions is the basis of desired quantitative analysis of medical images.
It is also very helpful for the research of various medical disorders.
Quantification of structural variation by accurate measurement of
volumes of region of interest can be used to evaluate severity of some
disease or evolution of some tissues.

Recently it has been widely accepted that deep neural networks (DNN) have an
impressive performance in various computer vision tasks.
 The techniques based on DNN have also been widely
 applied to the field of medical image segmentation and gain great
 success. The goal of segmentation with DNN is to allocate
 each pixel of the image with a corresponding category label. A lot
 of attempts have been made on dense pixel label prediction for
 medical images by developing various DNN. Brebisson
 et al.~[1] use the CNNs for anatomical brain segmentation
 and achieve better performance. Zhang et al.~[2] has
 proposed deep convolutional neural networks for extracting isointense
 stage brain tissues using multi-modality MR images.
 Li et al.~[3] apply the CNNs to extract the intrinsic image
  features of lung image patches.

 Long et al.~[4] developed the Fully Convolutional Networks (FCN),
 which is implemented based on VGG-16~[5].
 FCN is an end-to-end network, which can effectively solve the
 overstorage problem. He et al.~[6] proposed deep residual
  network  (ResNet), which efficiently  combine  information with
  extremely deep architectures and achieves compelling accuracy.
   Fisher et al.~[7] proposed dilated convolution, which
   can effectively enlarge receptive field without losing resolution.
   It improves the performance in VGG-16 network and accelerate
   convergence. Ronneberger et al.~[8] proposed U-Net
   for  biomedical image segmentation. U-Net can be trained end-to-end
   from very few images. The architecture of U-Net consists of a
   contracting  path to capture context and a symmetric expanding
   path that enables precise localization. Zhang  et al.~[9]
   proposed a pyramid dilated Res-U-Net based on ResNet
   and FCN with dilated residual unit. This net introduced
   LeakyReLU in the downsampling process and achieve desired
   performance for ultrasound nerve segmentation. Cui et al.~[10]
   proposed a deep network based on ResNet and U-Net, which
   connected to a fully-connected CRF to refine boundary
   information.The pyramid dilated convolution is designed to
   exploit global context features with multi-scale.

 Many great achievements have been made on various techniques of
 DNN for medical image segmentation. However, to our knowledge, so far the
 prior knowledge of the medical images has not been taken into consideration
 of the DNN-based approaches.
 Compared with natural images,
  medical images have a distinct feature: all images of the same
  imaging modality may contain same normal organs or tissues that
  are located at the similar positions in the images. This feature
  indicates that prior knowledge about organs or tissues is available
  in medical images, which can be utilized for coarse localization of
  these organs or tissues by registration methods. Most of current
  approaches based on deep nets train the kernel based on the
  information of whole images. Therefore the common features of
  similar but different structures may be learned in the medical
  images, such as the common features of left and right lungs in
  CT images. These features are useful for detecting and locating
  organs or tissues. But these global features may not be helpful
  for accurate segmentation of organs or tissues, which will be
  exemplified and analyzed in this paper.

 In order to make full use of prior knowledge for accurate
 segmentation of medical images, we proposed a knowledge-based
 fully convolutional network in this paper. In this new
 framework, the medical images are pre-partitioned into different
  regions based on prior knowledge, each of which contain different
  organs or tissues. A strategy of knowledge-based convolution
  kernels is introduced into our KFCN such that our framework
  consists of multiple channels equipped with different
  convolution kernals that correspond to different regions.
   Instead of using  information of whole image, in the KFCN model,
   each convolution kernel is trained with information only
   from the corresponding pre-partition region containing
   certain object. The difference between our strategy and
   current approaches is theoretically analyzed in
   section~\ref{S:Method and Theory}. Experiments about
   segmentation of lung CT images shows that our KFCN
   achieves better segmentation accuracy with the strategy
   of knowledge-based convolution kernels.

 \section{Method and Theory}\label{S:Method and Theory}
 Details of our knowledge based fully convolution network
 will be given in this section, along
 with some theoretical analysis of functions in KFCN and FCN.
 Some measurements necessary for our experiments to
 compare KFCN and FCN are given.

\subsection{Prior knowledge}
For most medical images, different objects tend to be organized
in a similar way, confined to a similar bounded region
in the image. For example, the left lung almost always
locates on the left
part of the image and the right lung on
the right part. In this way, we can usually
partition different objects
into different boxes.

For an image segmentation task, we use $X$ to stand for the image
variable. Then based on prior knowledge,
we can partition it to be $X=(X_1, X_2,...,X_n)$,
where $X_i$ is part of $X$ and mainly contains one object,
Also, we can segmentate this object in $X_i$ from the background
without knowing the contents of $X_j,~j\ne i$.

In traditional segmentation tasks,
fully convoluional network (FCN) is adopted and
in the convolutional layer of FCN, each convolutional kernel
operates on all $X_i$. In other words, this convolutional layer can
be formalized as
\begin{equation}
  f(X;W)=(X_1*W,X_2*W,...,X_n*W), \label{old-equ}
  \end{equation}
where $W$ is the convolutional kernel and $*$ represents the
convolution.

However, this kind of convolutional kernel
will be affected by all objects and their features in the image.
It will extract shared features
of different objects and may miss some of their
uniqueness, leading to poor segmentation performances.

In this paper, we attempt to incorporate prior knowledge of
images about different objects into the design of convolutional layer,
to construct convolutional kernels based on each kind of
object. Specifically, instead of using the function
in (\ref{old-equ}), we segmentate images with function
\begin{equation}
g(X;W)=(X_1*W_1,X_2*W_2,...,X_n*W_n)
\end{equation}
where $W$
could be seen as a kernel vector which contains
a couple of convolutional kernels $W_i,i=1,..,n$. In this way,
each convolutional kernel is designed and trained for a
specific object and feature. It turns out the kernel vector
works more professional, and leads to
better segmentation performances.

\subsection{Knowledge-based FCN}

Here we introduce the principle of constructing KFCN, for
a segmentation task and images $\{X^{(i)}\}$,
with the assumption that there
are $n$ different objects in each image.
At first, for image $X^{(i)}$, we apply a single shot
multibox detector (SSD [11]) to get
$n$ boxes $\{B^{(i)}_j,j=1,...,n\}$ in image each mainly
containing one object.
Then, we extract data in these boxes out and resize them
to have the same size, denoted as $\{Y^{(i)}_j,j=1,...,n\}$.
Next, we build $n$ independent small traditional FCN $N_j,j=1,...,n$
with $N_j$ only deals with $Y^{(i)}_j$ to get $n+1$ probability
maps $\{Z^{(i)}_{jk},k=1,..,n+1\}$, each represents the
probability of pixels in $Y^{(i)}_j$ belonging to
$n$ object class and the background class.
Then we resize $\{\{Z^{(i)}_{jk},k=1,..,n+1\},j=1,..,n\}$ back
to the size the original boxes $\{B^{(i)}_j,j=1,...,n\}$ to get
$\{\{U^{(i)}_{jk},k=1,..,n+1\},j=1,..,n\}$, each represents the
probability of pixels in each box belonging to $n$ object class
and the background class.

Then we make use an operation called sew up, for pixels in
$X^{(i)}$ which lie in a box, from above process, we have got
their probability map, for pixels outside of all boxes, we assume
they all belong to the background, and thus can get their
probability map, then we sew up these two parts together
to get the probability map $\{V^{(i)}_{k},k=1,..,n+1\}$
of $X^{(i)}$.

Figure 1 gives a specific structure of KFCN
which will be used in our following experiments
based on lung CT images.

\begin{figure}
  \centering
  \includegraphics[scale=0.25]{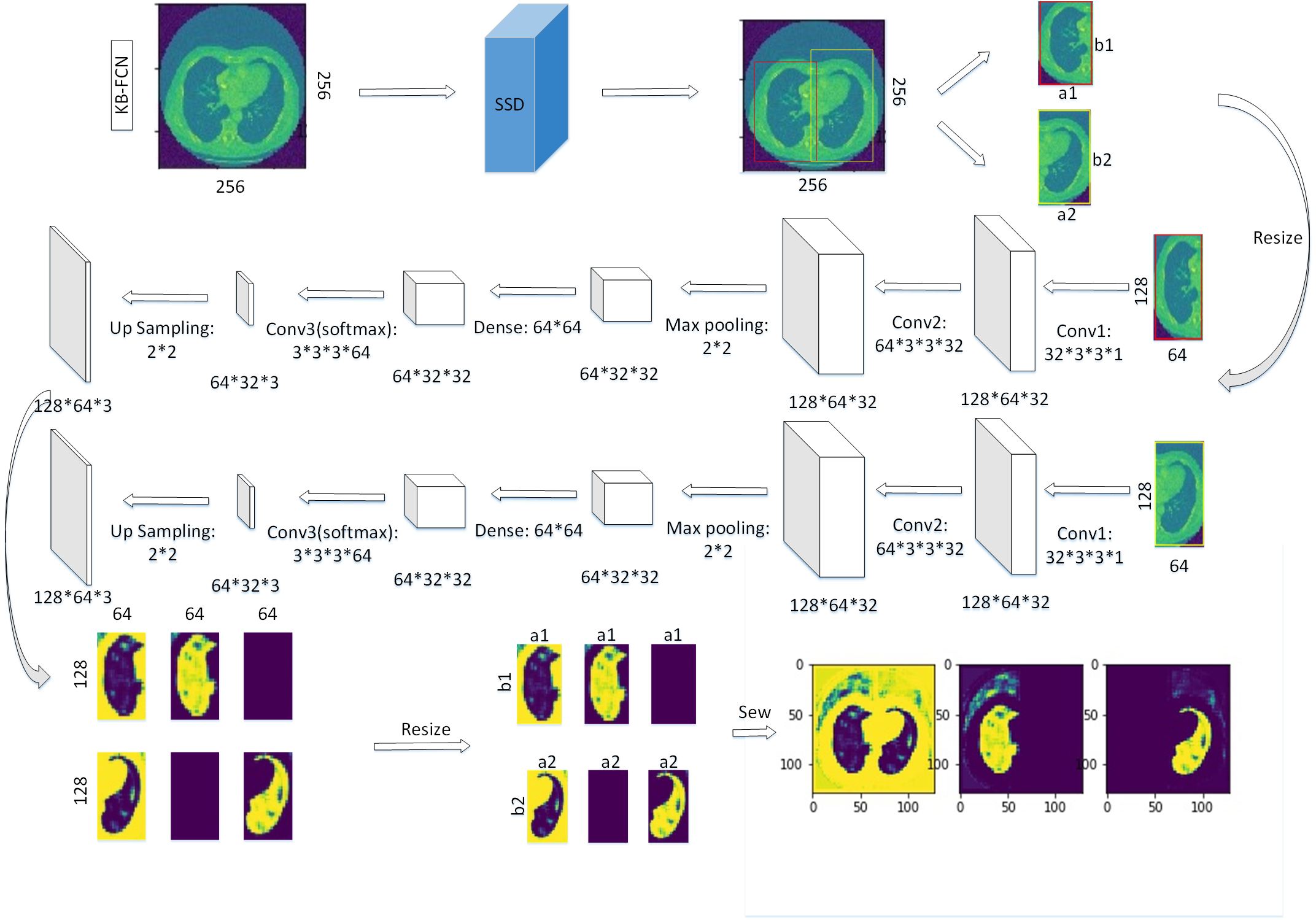}
  \label{stru}
  \caption{
  This is the structure of a KFCN used to segmentate lung CT images.
  On the first row, use a SSD to get the partition of images into
  boxes, then resize them into the fixed shape which were input to the
  next two channel, two small FCN in the second row.
  For each small FCN, it
  contains 6 layers, 3 convolution, 1 max pooling, 1 dense
  and 1 upsampling with specific size in the figure. In the third row,
  from two small FCN, we get probability maps and then resize them
  back to the size of original boxes, following a sew up operation
  to get the final probability maps of the image.}
\end{figure}

\subsection{Analysis of functions in KFCN and FCN}
In this subsection, we will evaluate the error of functions in
kb-fcn and fcn in a simple settings.
We consider only one convolutional layer based on ReLU as activation
function. Without loss of generality, we may flatten one image
variable $X$ to be a vector $\mathbf{x}$, where
$\mathbf{x}=(\mathbf{x_1},\mathbf{x_2})$
is a $(d_1+d_2)$ input data vector with
$\mathbf{x_1}$ of length $d_1$, $\mathbf{x_2}$ of length $d_2$ which is
 a partition of $\mathbf{x}$, let $u=g(\mathbf{x},\mathbf{t^*})
=(\mathbf{x_1}*\mathbf{t^*_1},\mathbf{x_2}*\mathbf{t^*_2})$
be the function which can accurately
segmentate images with desired $(d_3+d_4)$ output vector, where
$\mathbf{t^*}=(\mathbf{t^*_1},\mathbf{t^*_2})$ are the vector form
of convolutional kernels.

Then the function of KFCN
can be formalized as $v_1=g(\mathbf{x},\mathbf{t})=
(\mathbf{x_1}*\mathbf{t_1},\mathbf{x_2}*\mathbf{t_2})$, where
$\mathbf{t}=(\mathbf{t_1},\mathbf{t_2})$ are the vector form estimator
of convolutional kernels in KFCN. Comparatively the function of
traditional fcn can be formalized as
$v_2=g(\mathbf{x},\mathbf{t_{fc}})=
(\mathbf{x_1}*\mathbf{t_f},\mathbf{x_2}*\mathbf{t_f})$, where
$\mathbf{t_{fc}}=(\mathbf{t_f},\mathbf{t_f})$ are the vector form estimator
of convolutional kernels in FCN.

As a matter of fact, we can
write $\mathbf{x}*\mathbf{t}=\sigma(\mathbf{x}\mathbf{w})$, where
$\sigma$ is the RELU function and $\mathbf{w}$ is the
Toeplitz matrix of $\mathbf{t}$.
We regard $\mathbf{w}$ as the changing variable.
The next question would be to know that with $l_2$ loss
$E_1(\mathbf{w_1})=||u-v_1||^2$ and
$E_2(\mathbf{w_{fc}})=||u-v_2||^2$ (where $\mathbf{w_1},\mathbf{w_{fc}}$
are Toeplitz matrices of $\mathbf{t},\mathbf{t_{fc}}$), whether gradient
descent will converge to the desired solution
$\mathbf{w^*}=\mathbf{w_{t^*}}$. Note that
the gradient descent update is $\mathbf{w^{(l+1)}}=\mathbf{w^{(l)}}+\eta \Delta
\mathbf{w^{(l)}}$, where $\Delta \mathbf{w^{(l)}}=-\nabla E(\mathbf{w^{(l)}})$. Then,
let $\eta \to 0$, we get that $\dot{\mathbf{w}}=-\nabla E(\mathbf{w})$,
then $\dot{E}=-||\nabla E(\mathbf{w})||^2\leq 0$, the function
value $E$ is nonincreasing. Hence, we need to
check the points satisfying $\mathbf{w}\neq \mathbf{w^*}$ with
$\nabla E(\mathbf{w})=0$.

In our analysis below, we take the assumption that entries of input $\mathbf{x}$
follow Gaussian distribution.
In this situation, the gradient is a random variable
and $\Delta \mathbf{w} = -\mathbf{E} [\nabla E(\mathbf{w})]$.
The expected $\mathbf{E} [E(\mathbf{w})]$ is also
 nonincreasing no matter whether we follow
 the expected gradient or the gradient itself,
for $\mathbf{\dot{E}}
 = \mathbf{E} [-\nabla E(\mathbf{w})^T\nabla E(\mathbf{w})]\leq-\mathbf{E} [\nabla E(\mathbf{w})]^T\mathbf{E}
 [\nabla E(\mathbf{w})]\leq 0$.
Therefore, we analyze the behavior of expected gradient
$\mathbf{E} [\nabla E(\mathbf{w})]$ rather than $\nabla E(\mathbf{w})$.

For simplicity, let $d_1=d_2=d, d_3=d_4=1$, then we have following lemma.

\begin{lemma} (Lemma 3.1 in [12])
  \begin{equation}
\mathbf{E} [\Delta \mathbf{w_1}]=\frac{1}{2}(\mathbf{w}^*-\mathbf{w_1})+
\frac{1}{2\pi}
(\frac{\mathbf{w}^*}{||\mathbf{w}^*||}\sin{\theta}\mathbf{w_1}-\theta \mathbf{w}^*)\\
\end{equation}
\begin{equation}
\mathbf{E} [\Delta \mathbf{w}_{fc}]=\frac{1}{2}(\mathbf{w}^*_1-\mathbf{w}_{f}+
\mathbf{w}^*_2-\mathbf{w}_{f})+
\frac{1}{2\pi}
(\frac{\mathbf{w}^*_1}{||\mathbf{w}^*_1||}\sin{\theta_1}\mathbf{w}_{f}-
\theta_1 \mathbf{w}^*_1+
\frac{\mathbf{w}^*_2}{||\mathbf{w}^*_2||}\sin{\theta_2}\mathbf{w}_{f}-
\theta_2 \mathbf{w}^*_2)
\end{equation}
where $\theta,\theta_1,\theta_2\in [0,\pi]$ are the angles of
$\mathbf{w},\mathbf{w}^*$, $\mathbf{w}_{f}, \mathbf{w}^*_1$
and $\mathbf{w}_{f}, \mathbf{w}^*_2$.
\end{lemma}

Combined with following lemma, we can arrive a good conclusion about the
first function, which corresponds to the function in KFCN.
\begin{lemma} (Lemma 3.2 in [12])

When $\mathbf{w}_1^{(1)}\in \Omega=\{\mathbf{w}:||\mathbf{w}-\mathbf{w}^*||<
||\mathbf{w}^*||\}$, following
the dynamics of $\mathbf{E} [\Delta \mathbf{w}_1]$, the Lyapunov function
$V(\mathbf{w}_1)=\frac{1}{2}||\mathbf{w}_1-\mathbf{w}^*||^2$ has
$\dot{V}\leq 0$ the system
is asymptotically stable and thus $\mathbf{w}_1^{(t)}\to
\mathbf{w}^*$ when $t\to +\infty$
\end{lemma}

This ensures the convergence and correctness of our
model by making use of prior knowledge. However, we would like to see what's the
situation for
the second function corresponding to traditional FCN. Note that
\begin{equation}
\dot{V}=(\mathbf{w}_{f}-\mathbf{w}^*_1)^T\Delta \mathbf{w}_{f}
+(\mathbf{w}_{f}-\mathbf{w}^*_2)^T\Delta \mathbf{w}_{f}
\end{equation}
hence, we get $\dot{V}=-\mathbf{y}^TM\mathbf{y}$, where $\mathbf{y}=[||\mathbf{w}^*_1||,
||\mathbf{w}^*_2||,||\mathbf{w}_f||]^T$ and $M$ is the following 3-by-3 matrix:

\begin{multline} M=\frac{1}{4\pi}\left[\begin{array}{ccc}
\sin(2\theta_1)+2\pi-2\theta_1 & (2\pi-\theta_1-\theta_2)\cos(\alpha)+\sin(\theta_1+\theta_2)\\
(2\pi-\theta_1-\theta_2)\cos(\alpha)+\sin(\theta_1+\theta_2)&\sin(2\theta_2)+2\pi-2\theta_2 \\
-(2\pi-\theta_1)\cos(\theta_1)-\sin(\theta_1)&-(2\pi-\theta_2)\cos(\theta_2)-\sin(\theta_2)\\
\end{array}\right.\\
\left.\begin{array}{cc}
-(2\pi-\theta_1)\cos(\theta_1)-\sin(\theta_1)\\
-(2\pi-\theta_2)\cos(\theta_2)-\sin(\theta_2)\\
-4\pi\\
\end{array}\right]
\end{multline}

where $\alpha\in [0,\pi]$ is the angle between $ \mathbf{w}^*_1$ and
$\mathbf{w}^*_2$.

Then we consider the conditions $\theta_1,\theta_2$ should satisfy
so as to find an asymptotically stable region, that is, $M$ should be positive
definite in this region. It follows that all order principal minor determinant
$D_{11},D_{22},D_{33}$ should be positive. Because of
the complexity of the expression, we solve the equation numerically.
The result turns out that for any $\alpha\in [0,\pi]$, there is no region where $\theta_1,\theta_2\in [0,\pi]$
for $M$ to be positive definite.
Consequently, there is no asymptotically stable region in
this case.

Furthermore, when $d_3,d_4$ become larger, the system becomes complicated and
it is hard for us to analyze both cases, yet from theorem 4.1 in [12] ,
for the first function, we can still find an asymptotically stable region.
Hence, from this analysis, we see the advantage of KFCN, since
it represents a function with an asymptotically stable region, while traditional FCN
represents a function doesn't necessarily correspond to an asymptotically
stable region.

\subsection{Generalization ability and Similarity analysis}
We will check our assumption that convolutional kernels
 which were used to segmentate different objects with high accuracy
are different by considering the generalization ability of
convolutional kernels.
Generally speaking, we consider the performance of
convolutional kernels trained based on one object when they were used
to segmentate other objects.

Let $N_i, N_j$ be two small FCN embedded in KFCN,
for the generalition ability of convolutional kernels
in $N_i$, we set the values of convolutional kernels in
$N_j$ to be the values of convolutional kernels in
$N_i$, and then do segmentation again to see the results,
namely, segmentation performance, accuracy and loss.

By considering generalization ability of convolutional kernels,
we can directly see the different performances caused by
knowledge based convolutional kernels.
However, it remains to be shown how similar and different these
kernels are, which we will measure with the help of
point-wise similarity of two couples of convolutional kernels.

\begin{definition}
For two convolutional kernels $w_i$ and $w'_j$, their point-wise
similarity is defined as
\begin{equation}
Sim(w_i,w'_j)=\frac{w^v_i*w'^v_j}{||w^v_i||*||w'^v_j||}
\end{equation}
where $w^v_i, w'^v_j$ is the vector form of $w_i, w'_j$.
\end{definition}

\begin{definition}
Let $\{ w_i \}_{i=1}^{K}$ and $\{ w'_j \}_{j=1}^{K}$ be
two couples of convolutional kernels. Rearrange $\{ w'_j \}_{j=1}^{K}$
such that $Sim(w_i,w'_i)=\max_{j=1}^K Sim(w_i,w'_j)$.
Point-wise similarity of $\{ w_i \}_{i=1}^{K}$ with respect
to $\{ w'_j \}_{j=1}^{K}$ is defined as
\begin{equation}
Sim_l(\{ w_i \}_{i=1}^{K},\{ w'_j \}_{j=1}^{K})=\frac{1}{K}\sum_{i=1}^K Sim(w_i,w'_i)
\end{equation}
Similarly, point-wise similarity $Sim_r(\{ w_i \}_{i=1}^{K},\{ w'_j \}_{j=1}^{K})$ of $\{ w'_j \}_{j=1}^{K}$
with respect to $\{ w_i \}_{i=1}^{K}$ can be derived in the same way.
\end{definition}

\begin{definition}
Let $\{ w_i \}_{i=1}^{K}$ and $\{ w'_j \}_{j=1}^{K}$ be
two couples of convolutional kernels. Point-wise similarity
of $\{ w_i \}_{i=1}^{K}$ and $\{ w'_j \}_{j=1}^{K}$
is defined as
\begin{equation}
Sim(\{ w_i \}_{i=1}^{K},\{ w'_j \}_{j=1}^{K})=\frac{1}{2}(Sim_l+Sim_r)
\end{equation}

\end{definition}

We will calculate these values in our experiments so as to
measure the similarity of several couples of convolutional kernels.
\section{Experiments and Discussion}
\subsection{KFCN vs FCN}
\begin{figure}[H]
  \centering
  \subfigure[] {\includegraphics[scale=0.45]{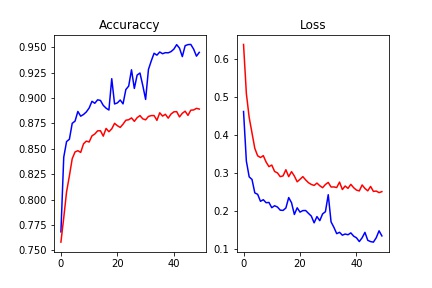}}
  \subfigure[] {\includegraphics[scale=0.45]{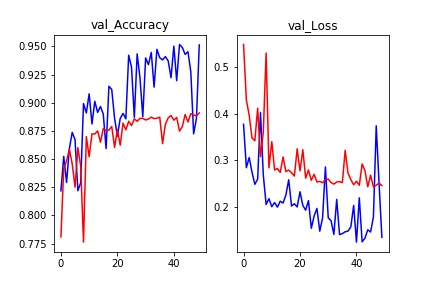}}
  \caption{The left part are curves in training set, the right
  part are curves in validation set. Blue curves represents KFCN while
  red curves represents baseline FCN.}
  \label{A-L-C}
\end{figure}

Based on a dataset on lung ct images with 246 images, we train
a KFCN of the structure described in Figure 1. The baseline
model is a traditional FCN with contains 6 layers, 3 convolution,
1 max pooling, 1 dense and 1 upsampling with the same size of the small FCN
in Figure 1, trained on the same dataset. Figures \ref{A-L-C} are the accuracy and loss curves of two networks
on training set and validation set during training.

Also, the accuracy and loss of two models on original
test set is shown in the left part of table \ref{A-L-T},
where the loss is calculated as mean squard error.
\begin{table}[H]
  \label{A-L-T}
  \caption{Accuracy and loss on test set}
  \centering
\begin{tabular}{c|c|c|c|c}
\toprule
Image &\multicolumn{2}{|c|}{Original images}&\multicolumn{2}{|c}{Flip images}\\
\midrule
Property &Accuracy &Loss &Accuracy &Loss \\
\midrule
KFCN  &  $95.6\%$ &0.022 & $77.5\%$ & 0.125\\
\midrule
FCN   & $89.1\%$ & 0.048 & $80.6\%$ & 0.129 \\
\bottomrule
\end{tabular}
\end{table}

\begin{figure}[H]
  \centering
  \includegraphics[scale=0.8]{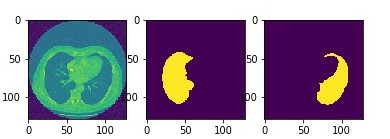}
  \caption{Original lung CT image and its ground truth}
  \label{O-T}
\end{figure}

Moreover, for an example of lung CT image, shown in Figure \ref{O-T},
where the left image is the original training image, the
middle image is the ground truth of left lung and the right image
is the ground truth of the right lung.

\begin{figure}[H]
  \centering
  \includegraphics[scale=1]{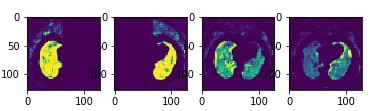}
  \caption{Segmentatation of images by KFCN and FCN}
  \label{M-O-T}
\end{figure}

we show the corresponding segmentation by KFCN and traditional FCN
respectively in figure \ref{M-O-T}, the first 2 images were segmentation of left
lung and right lung by KFCN and the last 2 images were segmentation by
traditional FCN.

As you can see, KFCN and FCN are segmentating lungs in a different
way. With knowledge based convolutional kernels, KFCN can
extract different features and achieve a remarkable accuracy.
On the other hand, though traditional FCN achieves a pretty high
accuracy, its segmentations are not accurate and reasonable actually.
Convolutional kernels in traditional FCN were affected by both
features and can not distingunish different features well,
which leads to a bad performance.

\subsection{Generalization ability}
Now we turn to consider the generalition ability of different
knowledge based convolutional kernels.
For generalization ability of convolutional kernels trained
only on the left lung in KFCN, we apply them to do
segmentation of the right lung. Conversely, the generalization
ability of convolutional kernels trained only on the right lung in KFCN.
Similarly, the generalization ability of convolutional kernels
in FCN is taken into account. Actually, for lung CT images, we can just flip
the left part and right part to get the same effect.

\begin{figure}[H]
  \centering
  \includegraphics[scale=1]{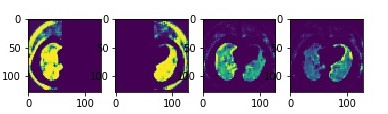}
  \caption{The first image is the segmentation when
 right kernels in KFCN were used to segmentate the left lung, the second
 image is the segmentation when left kernels in KFCN were used to segmentate
 the right lung. The last two images are the segmentation
 of kernels in FCN to the same flip image.}
  \label{M-F-T}
\end{figure}

The segmentation of flip image of figure3 by two models is shown in
Figure \ref{M-F-T}.

In this case, accuray and loss of KFCN and FCN were shown in the
right part of table \ref{A-L-T}. It can be seen that convolutional kernels in
KFCN have poor generality ability since they are specialized
according to different objects, finding different features.
However, convolutional kernels in traditional FCN have good generalition
 ability, extracting features shared by different
objects. As a matter of fact, since these kernels operate on
all image, so the segmentations vary a little compared
to the segmentation of original image. So,
These convolutional kernels which are to segmentate different
objects with high accuracy are different to some extent while
traditional FCN will ignore this prior knowledge.

\subsection{Similarity analysis}
We now have three couples of convolutional kernels with the same size,
namely, $W_L=\{w_{l_i}\}_{i=1}^{n}$, convolutional kernels trained
in KFCN operating on the left lung, $W_R=\{w_{r_i}\}_{i=1}^{n}$,
convolutional kernels
trained in KFCN operating on the right lung,
$W_T=\{w_{t_i}\}_{i=1}^{n}$, convolutional kernels
trained in traditional FCN operating on the whole image.
The point wise similarity of them is calculated in table \ref{pt-s}.

\begin{table}[H]
  \caption{Point-wise similarity}
  \label{pt-s}
  \centering
  \begin{tabular}{l|llll}
    \toprule
    layer     & kernels     & $Sim_l$     & $Sim_r$ & $Sim$\\
    \midrule
    \multirow{3}*{$C_1$}&$W_L,W_R$&0.6908&0.6902&0.6905\\
    & $W_L,W_T$ & 0.8633 & 0.8518  & 0.8576    \\
    & $W_R,W_T$ & 0.6630 & 0.6885  & 0.6758  \\
    \midrule
    \multirow{3}*{$C_2$}&$W_L,W_R$&0.2550&0.2548&0.2549\\
    & $W_L,W_T$ & 0.2805 & 0.2757  & 0.2781    \\
    & $W_R,W_T$ & 0.2395 & 0.2384  & 0.2390  \\
    \midrule
    \multirow{3}*{$C_3$}&$W_L,W_R$&0.1536&0.1428&0.1482\\
    & $W_L,W_T$ & 0.1115 & 0.0865  & 0.0990    \\
    & $W_R,W_T$ & 0.0692 & 0.1054  & 0.0873  \\
    \bottomrule
  \end{tabular}
\end{table}

It can be seen that only on the first layer, these kernels are similar,
in the following layers, convolutional kernels vary a lot
and have little in common.
As a matter of fact, in the first layer, both networks are extracting
similar basic features, what's more, convolutional kernels
in traditional FCN are similar to the
convolutional kernels in KFCN in the first layer, yet in the following layers,
these features are organized differently depends on different objects.

\section{Conclusion}
In this paper, we proposed a novel framework KFCN for medical
image segmentation by incorporating prior knowledge into the design
and training of convolutional kernels.
 Based on theoretical analysis of functions and error in KFCN and
traditional FCN, we conclude that KFCN has an asymptotically stable region while
traditional FCN does not necessarily do.
This discovery demonstrates the convergence superiority
of KFCN, which can converge asymptotically under mild conditions.
Meanwhile, experimental results show that
KFCN outperforms traditional FCN in segmentation of lung CT images.
Furthermore, the numerical results about difference
 among convolutional kernels in terms of generalization ability and
 point-wise similarity, validates our assumption about incorporation
 of prior knowledge into KFCN, and illustrates the advantages of our
 knowledge-based convolutional kernels. In the future research, we will
 test KFCN for segmentation of various medical images and try to
 introduce KFCN framework into semantic segmentation of natural images.
Another interesting research topic is to see the number of images required
for KFCN to achieve a high accuracy.

\section*{References}
\small
[1] Brebisson, A.D.\ \& Mountana, G.\ (2015) Deep neural
networks
for anatomical brain segmentation. {\it Proceedings of the IEEE Conference
on Computer Vision and Pattern Recognition Workshops}.

[2] Zhang, W. Li, R., Deng, H.\ \& Wang, L.\ (2015) Deep
convolutional neural networks for multi-modality isointense infant
brain image segmentation. {\it NeuroImage}{\bf 108}, pp. \ 214-224.

[3] Li, Q., Cai, T., Wang, X., Zhou, Y.\ \& Feng, D.\ (2014)
Medical
image classification with convolutional neural network. {\it the 13th
International Conference on Control Automation Robotics \& Vision
(ICARCV). IEEE}

[4] Long, J., Shelhamer, E.\ \& Darrell, T.\ (2015)
Fully
convolutional networks for semantic segmentation. {\it Proceedings
of the IEEE Conference on Computer Vision and Pattern Recognition}

[5] Simonyan, K.\ \& Zisserman, A.\ (2014)
Very deep convolutional
networks for large-scale image recognition[J]. {\it ArXiv preprint
arXiv:1409.1556}

[6] He, K., Zhang, X., Ren, S.\ \& Sun, J.\ (2016)
Deep residual
learning for image recognition. {\it Proceedings of the
Institute of Electrical and Electronics Engineers Conference
on Computer Vision and Pattern Recognition}

[7] Yu, F.\ \& Koltun, V.\ (2015) Multi-scale context
aggregation by dilated convolutions. {\it ArXiv preprint
arXiv:1511.07122}

[8] Ronneberger, O., Fischer, P.\ \& Brox, T.\ (2015)
U-net: Convolutional networks for biomedical image segmentation[C].
 {\it International Conference on Medical Image Computing and
 Computer-Assisted Intervention.} pp.\ 234-241. Springer, Cham.

[9] Zhang, Q., Cui, Z., Niu, X., Geng, S.\ \& Qiao, Y.\ (2017)
 Image Segmentation with Pyramid Dilated Convolution based on
 ResNet and U-Net. {\it 24th International Conference on Neural
 Information Processing}

[10] Cui, Z., Zhang, Q., Geng, S., Niu, X., Yang, J.\ \&
Qiao, Y.\ (2017) Semantic Segmentation with Multi-path Refinement and
Pyramid Pooling Dilated-Resnet. {\it 2017 IEEE International Conference
on Image Processing}

[11] Liu, W., Anguelov2, D., Erhan3, D., Szegedy3, C.,
Reed4, S., Fu, C., Alexander C. Berg1\ (2016)
SSD: Single Shot MultiBox Detector.
{\it European Conference on Computer Vision}

[12] Tian, Y.\ (2017) Symmetry-breaking convergence
analysis of certain two-layered neural networks
with RELU nonlinearity. {\it International Conference on Learning Representations}

\end{document}